\documentclass[]{ceurart}
\usepackage{listings}
\usepackage[obeyFinal]{todonotes}

\usepackage[T1]{fontenc}
\usepackage{lipsum}
\usepackage{tabulary}
\usepackage{lscape}
\usepackage{verbatim}
\usepackage{graphicx}
\usepackage{subcaption}
\usepackage{comment}
\usepackage{soul}
\usepackage{siunitx}
\usepackage{algorithmicx}%
\usepackage{algorithm}%
\usepackage{algpseudocode}

\usepackage{soul}

\usepackage{mathptmx}
\usepackage{comment}
\usepackage{array}
\usepackage{ragged2e}
\usepackage{amsmath,amssymb,amsfonts}%

\newcommand{\tuple}[1]{\langle#1\rangle}

\newcommand{{\model}}{\emph{OPM}}
\newcommand{{\moe}}{\emph{MoE-{\em \model}}}
\newcommand{{\moeName}}{\emph{Mixture-of-Experts-{\em \model}}}
\newcommand{{\algo}}{\emph{MoE-{\em \model} Discovery}}
\newcommand{\nop}[1]{}
 
\newcommand{\N}{\mathcal{N}}

\newcommand{\regweight}{\lambda_R}

\begin{document}
\copyrightyear{2024}
\copyrightclause{Copyright for this paper by its authors.
Use permitted under Creative Commons License Attribution 4.0 International (CC BY 4.0).}
\conference{Proceedings of the 1st International Workshop on Explainable Knowledge Aware Process Intelligence, June 20--22, 2024, Roccella Jonica, Italy}

\title{Discussion: Effective and Interpretable Outcome Prediction by Training Sparse Mixtures of Linear Experts}
\tnotemark[1]
\tnotetext[1]{
This paper summarizes results presented at workshop \emph{ML4PM 2023}, associated with conference ICPM 2023, October 23-27, 2023, Rome, Italy, and published in~\cite{folino2024moe}.
}

\author[1]{Francesco Folino}[%
email=francesco.folino@icar.cnr.it
]
\author[1]{Luigi Pontieri}[%
email=luigi.pontieri@icar.cnr.it
]
\cormark[1]
\author[1]{Pietro Sabatino}[%
email=pietro.sabatino@icar.cnr.it 
]
\address[1]{Institute for High Performance Computing and Networking (ICAR-CNR), via P. Bucci 8/9C, 87036 Rende (CS), Italy }

\cortext[1]{Corresponding author.}

\begin{abstract}
Process Outcome Prediction entails predicting a discrete property of an unfinished process instance from its partial trace. High-capacity outcome predictors discovered with ensemble and deep learning methods have been shown to achieve top accuracy performances, but they suffer from a lack of transparency. Aligning with recent efforts to learn inherently interpretable outcome predictors, we propose to train a sparse Mixture-of-Experts where both the ``gate'' and ``expert'' sub-nets are Logistic Regressors. 
This ensemble-like model is trained end-to-end while automatically selecting a subset of input features in each sub-net, as an alternative to  the common approach of performing a global feature selection step prior to model training. 
Test results on benchmark logs confirmed the validity and efficacy of this approach. 
\end{abstract}

\begin{keywords}
  Process Mining \sep 
  Machine Learning \sep 
  XAI 
\end{keywords}

\maketitle

\section{Introduction}
\emph{(Process) Outcome Prediction} problem~\cite{Teinemaa2019} refers to the problem of predicting the outcome of an unfinished process instance, based on its associated \emph{prefix trace} (i.e., the partial sequence of events available for the instance).
Recently, different supervised learning approaches to this problem were proposed, which allow for discovering an outcome prediction model from labeled traces.
Outstanding performances in terms of prediction accuracy have been achieved by big ensembles of decision rules/trees discovered with random forest or gradient boosting algorithms~\cite{Teinemaa2019}, and \emph{Deep Neural Networks} (\emph{DNNs})~\cite{rama2021deep}.
However, the approximation power of these models comes at the cost of an opaque decision logic, which makes them unfit for settings where explainable predictions and interpretable predictors are required.
The call for transparent outcome prediction first originated several proposals relying model-agnostic post-hoc explanation methods~\cite{galanti2020_shap4pm,Rizzi2020ExplainabilityIP}
or explanation-friendly DNN-oriented solutions~\cite{wickramanayake2022_interpretablebpm,stierle2021_relevancescores}.

Due to widespread concerns on the reliability of attention and post-hoc attribution-based explanations
~\cite{slack2021reliable}, 
the discovery of inherently-interpretable  outcome predictors~\cite{stevens2022_xaiooppm,pcam2021_fox} was proposed of late.
Two alternative kinds of interpretable models, both leveraging \emph{Logistic Regression} (\emph{LR}) models as a building block, were exploited in~\cite{stevens2022_xaiooppm} to discover  outcome predictors from flattened log traces: 
\emph{Logit Leaf Model} (\emph{LLM}), a sort of decision tree where each leaf hosts an LR sub-model; and (ii) 
and \emph{Generalized Logistic Rule Model} (\emph{GLRM}),
where a single LR model is built upon the original feature and novel features, defined as conjunctive rules over subsets of the former and derived via column generation. 
These kinds of models were both shown to improve plain LR predictors by capturing some non-linear input-output dependencies.
An approach leveraging a neural implementation of fuzzy logic, named \emph{FOX}, was proposed in~\cite{pcam2021_fox} to extract easy-to-interpret IF-THEN outcome-prediction rules, each of which contains a fuzzy set for each (flattened) trace feature and a membership score for each outcome class. 
Unfortunately, as observed~\cite{stevens2022_xaiooppm} and discussed in Section~\ref{sec:complexity}, these LLM-based and GLRM-based methods may return cumbersome models/rules that hardly enable a clear and complete understanding of the predictor's behavior, 
On the other hand, if the global feature selection and (3-way) feature binning of FOX~\cite{pcam2021_fox} helps control prediction rules' size, it risks causing some information/accuracy loss.

We next describe an approach to outcome prediction that builds an ensemble of LR models up by training a Mixture of Experts (MoE) neural net. 
This net consists of multiple ``experts'' (specialized outcome predictors) and a sparse ``gate'' module devoted to routing each data instance to one of the experts. 
For the sake of interpretability, both the gate and experts take the form of LR classifiers (i.e. one-layer neural net with linear activations).
Differently from~\cite{imoe}, the user is allowed to control the complexity of the model by fixing the maximum number $kTop$ of features that the gate and each expert sub-net can use and the maximum number $m$ of experts. 
Instead of preliminary performing a global feature selection step as in~\cite{stevens2022_xaiooppm} and~\cite{pcam2021_fox}, first the neural net is trained using all the data features, and then the less important learned parameters are pruned out in a ``feature-based'' way. 
This allows different experts to use different subsets of the input features when making their predictions.

\section{Proposed Outcome-Prediction Model and Training Algorithm}
A \emph{Mixture of Experts}  (\emph{MoE}) is a neural net that implements both the gate and the local predictors (``experts'') through the composition of smaller interconnected sub-nets.
In particular, in classical (dense) MoEs~\cite{moe},
once provided with an input data vector $x$, the gate is a feed-forward sub-net that computes a vector of (softmax-ed) weights, one for each expert, estimating how competent they are in making a prediction for $x$.
The overall prediction for $M(x)$ is obtained by linearly combining all the experts' predictions for $x$ according to competency weights returned by the gate. In Sparse MoEs~\cite{shazeer2017outrageously}, this formulation is adapted to activate only a given number $k$ of experts, selected as those that were assigned the  top-\( k \) competency weights, for $x$, by the gate.

\medskip \noindent 
\textbf{Model $\model$} 
The outcome-prediction neural-network model proposed in our approach, named $\moeName$, can be regarded as a tuple $\N = \tuple{\N_g,\N_1,\ldots,\N_m}$  where $\N_E$ is the sub-net consisting of all its experts $\N_1, \ldots, \N_m$, and $\N_g$ is the gate sub-net.
This model as a whole encodes a function $f: \mathbb{R}^d \rightarrow [0,1]$ defined as follows:  
$f(x)  \triangleq \N_k(x)$ such that $k = \arg\max_{k \in \{1,\ldots,m\}} \N_g(x)[k]$ and $\N_g(x)[k]$ is the $k$-th component of the probability vector returned by the gate sub-net $\N_g$ when applied to $x$.

Thus, for any novel input instance $x$, a decision mechanism is applied to the output of the gate, which transforms it into an ``argmax''-like weight vector where all the entries are zeroed but the one corresponding to the expert which received the the highest competency score (which is turned into 1); this makes the gate implement a hard-selection mechanism.

For the sake of interpretability, the following design choices are taken:
\emph{(i)} each expert $\N_i$, for $i \in [1..m]$ is implemented as one-layer feed-forward nets with linear activation functions, followed by a standard sigmoid transformation:
\emph{(ii)} the gate $\N_g$ is implemented as a one-layer feed-forward network with linear activation functions followed by a softmax normalization layer. 

\medskip \noindent 
\textbf{Training algorithm:$\ \algo$}
The proposed training algorithm, named $\algo$, takes two auxiliary arguments as input: the desired number $m$ of expert sub-nets and the maximal number $kTop \in \mathbb{N} \cup \{\texttt{ALL}\}$ of input features per gate/expert sub-net (where \texttt{ALL} means that no actual upper bound is fixed for the latter number).

The algorithm performs four main steps:
\textbf{(1)} 
A $\moeName$ instance $M$ is created, according to the chosen number $m$ of experts, and initialized randomly. 
\textbf{(2)} 
$M$ is trained end-to-end using an batch-based SGD-like optimization procedure (using different learning rates for the gate and the experts and a variant of the loss function proposed in~\cite{moe}, favouring expert specialization and competency weights' skeweness).
\textbf{(3)} 
$M$ is optimized again with the same procedure but keeping the expert sub-nets frozen, to fine-tune the gate one only.
\textbf{(4)} 
Feature-wise parameter pruning is performed on both the gate and experts to make all these LR-like sub-nets  base their predictions on $kTop$ data features at most.
The loss function utilized in the training algorithm combines an accuracy term alike the one proposed in~\cite{moe} (favoring expert specialization) with a  regularization term summing up the absolute values of all the model parameters. 
The influence of this regularization term can be controlled via weighting factor $\lambda_R$.

The last step of the algorithm consists in applying an ad hoc, magnitude-based, structured parameter pruning procedure to both the gate $\N_g$ and all experts $\N_1, \ldots, \N_m$. In this procedure, each parameter block gathers the weights of the connections reachable from a distinct input neuron, and all the parameters that do not belong to any of $kTop$ blocks are eventually zeroed. This corresponds to making all the sub-nets $\N_g$, $\N_1, \ldots, \N_m$ to only rely on $kTop$ input features.

\section{Experiments}\label{sec:experiments}
Algorithm \algo was tested against dataset extracted from the
benchmark log \emph{BPIC 2011} and \emph{Sepsis}, 
obtained by making each prefix trace undergo the aggregation encoding after extending them all with timestamp-derived temporal features (e.g., weekday, hour, etc.).
Each dataset was partitioned into training, validation, and test sets by using the same 80\%-20\%-20\% temporal split as in~\cite{pcam2021_fox,Teinemaa2019,stevens2022_xaiooppm}.
The accuracy of each discovered model was evaluated by computing the AUC score for all test prefixes containing at least two events, as done in~\cite{pcam2021_fox,Teinemaa2019,stevens2022_xaiooppm}. 

Algorithm $\algo$ was using 100 epochs in both the end-to-end training and in the gate fine-tuning steps, without performing any post-pruning training (for the sake of efficiency). The number $m$ of experts was fixed to 6 empirically (after trying several values in [$2, \ldots, 16$]), since this choice ensured a good accuracy-vs-simplicity trade-off. 
Different configurations were tested instead for hyperparameter $\regweight$ (from 0.1 to 0.6) and $kTop$ (from 2 to 8).
Details on the setting of other parameters (e.g., batch sizes, learning rates can be found in~\cite{folino2024moe}.

As terms of comparison, we considered state-of-the-art outcome-prediction methods \emph{FOX}~\cite{pcam2021_fox} and \emph{GLRM}~\cite{stevens2022_xaiooppm},
and a baseline method, denoted as \emph{1-LR}, that discovers a single LR model ---the latter was simulated by running Algorithm $\algo$ with $m=1$ and $kTop=ALL$. 

\medskip \noindent 
\textbf{Prediction accuracy results}
Table~\ref{tab:comparison_table} reports the AUC scores obtained by the 6-expert $\moe$ models. 
Notably, $\algo$ often outperforms the baseline \emph{1-LR} in different $kTop\neq \texttt{ALL}$ configurations over some datasets, namely \texttt{bpic2011\_1}, \texttt{bpic2011\_4}, \texttt{sepsis\_1} . For the remaining datasets, there is always at least one $kTop$ configuration where $\algo$ performs better than the baseline,  when combined with feature selection. 
In particular, on average, $\algo$ achieves an AUC improvement of more than $20\%$ over \emph{1-LR}, with peaks reaching beyond $80\%$. 
This confirms that training multiple local LR outcome predictors usually improves the performance of training a single LR model on all the data features (as done by \emph{1-LR}).
In addition, $\algo$ always surpasses state-of-the-art methods FOX and GLRM on all the datasets but \texttt{bpic2011\_2} and \texttt{bpic2011\_3}, where some of them perform as well as $\algo$.

\begin{table}[t]
\centering
\caption{AUC scores obtained by: $\algo$ (run with $m=6$ and several values of $kTop$), the baseline method \emph{1-LR} and two competitors. For each dataset, the best score is shown in \textbf{\underline{Bold and underlined}}; $\algo$'s scores are shown in \textbf{Bold} when it outperforms the competitors.
}
\label{tab:comparison_table}
\small
\begin{tabular}{l | l | l | l | l | l|l|l|l}
\hline
\multirow[t]{3}{*}{Dataset} & \multicolumn{5}{l|}{\algo} & \multicolumn{3}{l}{Competitors}\\
\cline{2-9}
& \multicolumn{5}{l|}{\it kTop} & \emph{1-LR} & {\it FOX \cite{pcam2021_fox} } &{\it GLRM \cite{stevens2022_xaiooppm}}\\ 
\cline{2-6}
& 2 & 4 & 6 & 8 & ALL & & & \\
\hline
\texttt{bpic2011\_1}& 0.97& 0.95& 0.96& \bf  \underline{0.98}&  0.88 & 0.94& 0.97 & 0.92\\
\hline
\texttt{bpic2011\_2}& 0.85& 0.84& 0.86& \bf \underline{0.97}&  0.87 & 0.94& 0.92& \bf \underline{0.97}\\
\hline
\texttt{bpic2011\_3}& 0.95& \bf  \underline{0.98}& 0.96& \bf  \underline{0.98}& 0.91 & 0.97& \bf  \underline{0.98}& \bf  \underline{0.98}\\
\hline
\texttt{bpic2011\_4}& 0.69& 0.81& 0.80& 0.81& 0.80 & 0.68& \bf \underline{0.89}& 0.81\\
\hline
\texttt{sepsis\_1}& 0.49& 0.55& 0.56& \bf \underline{0.58}&  0.49 & 0.47& \bf \underline{0.58}& 0.47\\
\hline
\texttt{sepsis\_2}& 0.56& 0.56& \bf \underline{0.75}& 0.73&  0.72 & 0.74& 0.73& 0.73\\
\hline
\texttt{sepsis\_3}& 0.56& 0.61& \bf \underline{0.72}& \bf \underline{0.72}& 0.69 & 0.70& 0.68& 0.65\\
\hline
\end{tabular}
\end{table}

Generally, $\algo$ seems to perform worse when using very few features  than when trained with a slightly larger feature set (namely, $kTop=6,8$). 
However, on dataset \texttt{bpic2011\_1}, $\algo$ manages to achieve outstanding AUC scores even when using just two (resp. four) features. 
The compelling AUC results obtained by the proposed approach using less than 9 input features per sub-model 
provides some empirical evidence of its ability to support the discovery of more compact outcome predictors and easier-to-interpret prediction explanations compared to the state-of-the-art method FOX, as discussed below.

\medskip \noindent 
\textbf{Model/explanation complexity}\label{sec:complexity}
Generally, the lower the description complexity of a prediction model, the easier to interpret it and its predictions. 
In the cases of $\algo$ and baseline \emph{1-LR} this complexity can be computed by counting the non-zero parameters appearing in the respective LR (sub-)models, 
while the complexity of FOX model is the number of conditions appearing in its fuzzy rules.
When applied to the pre-filtered datasets \texttt{bpic2011\_1}, \texttt{bpic2011\_1}, $\ldots$, \texttt{bpic2011\_4}, \texttt{sepsis\_1}, $\ldots$, \texttt{sepsis\_3} (containing 4, 7, 6, 2, 5, 4 and 6 data features, respectively), FOX finds a model consisting of
81, 2187, 729, 9, 243,  81 and 729 fuzzy rules~\cite{pcam2021_fox}, respectively. 
Thus, the complexity of these FOX models 
range from 18 to 15309.

\begin{figure}[t]
\centering
\includegraphics[width=0.53\textwidth]{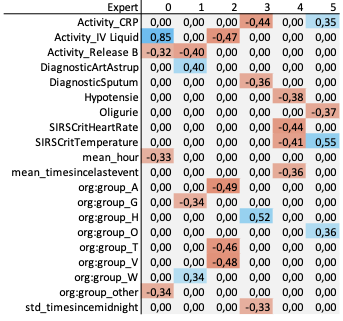}
\caption{Example $\moe$ discovered by $\algo$ (with $kTop=4$) from dataset \texttt{sepsis\_3}: parameter weights of the six LR experts. 
}
\label{figure:weights_explanation}
\end{figure}

\medskip \noindent 
\textbf{Qualitative results: an example of discovered $\moe$}\label{sec:qualitative}
Figure~\ref{figure:weights_explanation} shows the input features and associated weights that are employed by the six LR experts discovered when running algorithm $\algo$ with $kTop=4$ on dataset \texttt{sepsis\_3}, for which the outcome-prediction task is meant to estimate the probability that an in-treatment patient will leave the hospital with the prevalent release type (i.e., `Release A'). 
Only $20$ of the $86$ data features are used by the experts in total, but the specific feature subset of the experts differ appreciably from one another. 
In a sense, this means that the experts learned different input-output mappings (capturing different contex-dependent process-outcome use cases).

For instance, in predicting class 1, Expert 0 attributes a positive influence to `Activity\_IV Liquid' 
and negative influence to `Activity\_Release B' (a specific discharge type), `mean\_hour', 
and `org:group\_other' (a hospital group). 
Expert 1 attributes instead a positive influence to `DiagnosticArtAstrup' (arterial blood gas measurement) and `org:group\_W' and negative influence to both `Activity\_Release B' and `org:group\-\_G'. 
Analogous interpretations can be extracted from the remaining expert models, which also focus on specific activities and hospital groups. 

\section{Discussion and Conclusion}
The experimental results presented above confirm that the models discovered by $\algo$ exhibit a compelling trade-off between the accuracy and explainability of the their outcome predictions. 
This descends from both the modularity and conditional-computation nature of $\model$ models (where only one specific expert is chosen to make a prediction), and from the possibility to control the complexity of both the model and of their explanations (via hyperparameters $m$ and $kTop$). 
In particular, by focus on a small number of feature importance scores, the used is allowed to easily inspect and assess the internal decision logic of the model and get simple, faithful explanations for its predictions. 

Interesting directions of future work are: (i) converting LR-like sub-models returned by $\algo$ into logic rules, (ii) tuning hyper-parameters $kTop$ and $m$ automatically; (iii) evaluating the relevance of $\model$s' explanations through user studies.

\bibliography{biblio}

\end{document}